\documentclass[conference]{IEEEtran}
\IEEEoverridecommandlockouts
\usepackage{preamble}
\begin{document}

\title{Test Bench Study on Attitude Estimation\\ in Ground Effect Region Based on Motor Current\\for In-Flight Inductive Power Transfer of Drones
}
\author{\IEEEauthorblockN{Kota Fujimoto$^1$, Sakahisa Nagai$^2$, Nguyen Binh Minh$^3$, Hiroshi Fujimoto$^4$}
\IEEEauthorblockA{\textit{Graduate School of Engineering} \\
\textit{The University of Tokyo}\\
Kashiwa, Chiba, Japan \\s
fujimoto.kota21@ae.k.u-tokyo.ac.jp$^1$, (nagai-saka$^2$, nguyen.binhminh$^3$)@edu.k.u-tokyo.ac.jp, fujimoto@k.u-tokyo.ac.jp$^4$}
}

\maketitle
\thispagestyle{titlepagestyle} 

\begin{abstract}
    To overcome the short flight duration of drones, research on in-flight inductive power transfer has been recognized as an essential solution.
    Thus, it is important to accurately estimate and control the attitude of the drones which operate close to the charging surface.
    To this end, this paper proposes an attitude estimation method based solely on the motor current for precision flight control in the ground effect region.
    The model for the estimation is derived based on the motor equation when it rotates at a constant rotational speed.
    The proposed method is verified on the simulations and experiments.
    It allows simultaneous estimation of altitude and pitch angle with the accuracy of 0.30$\hspace{0.5mm}$m and 0.04$\hspace{0.5mm}$rad, respectively.
    The minimum transmission efficiency of the in-flight power transfer system based on the proposed estimation is calculated as 95.3$\hspace{0.5mm}$\%, which is sufficient for the efficient system.
\end{abstract}


\section{Introduction}
\begin{figure}[tb]
        \centering
        \includegraphics[width=0.9\linewidth]{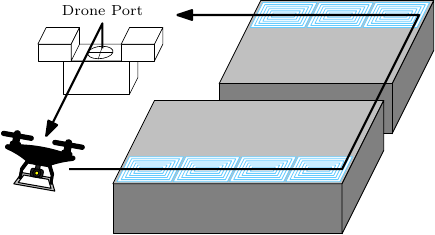}
        \caption{Security mission with in-flight inductive power transfer.}
        \label{fig:drone use case on security mission}
    \end{figure}

    Drones have been increasingly utilized in many aspects of human society.
    There are many promising applications where the drones have to operate in the desirable route.
    A typical application is security surveillance with the mission schematic shown in \cref{fig:drone use case on security mission}.
    In the diagram, charging the battery every lap is needed, which deteriorates the drone's operating rate.
    To realize the continuous monitoring capability, a lot of drones are required; however, this increases system cost a lot to assign enough drones to complete the mission.
    \par
\begin{figure}[tb]
        \centering
        \includegraphics[width=0.6\linewidth]{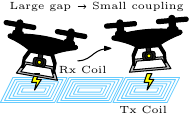}
        \caption{Deterioration of the coupling coefficient $k$ without precise attitude control on in-flight inductive power transfer system.}
        \label{fig:image of in-flight charging drone}
    \end{figure}

    Although much research has been conducted to extend the flight duration, an inductive power transfer for flying drones \cite{Chen2023-va,Arteaga2019-le} is the most effective way to increase the operating rate in fixed route operations. 
    The diagram of the in-flight inductive power transfer system is shown in \cref{fig:image of in-flight charging drone}. 
    In this system, the drones fly near the transfer coils placed on the roof or wall, and they are wirelessly powered while monitoring the area.
    \par
    This system makes it challenging to avoid power deviation and efficiency deterioration because drones fly along the transmitter coils while fluctuating, as shown in \cref{fig:image of in-flight charging drone}. 
    Previous studies realize the constant power transfer by controlling the converters \cite{Zhang2022-vq,Gu2022-cu}.
    However, they assumed that the mutual inductance fluctuates, so does not consider improving efficiency just by suppressing the mutual inductance deviation.
    \par
    To maintain the stability and high efficiency of in-flight power transfer, it is essential to precisely estimate and control both the altitude and attitude of the drone body.
    This would be great challenge due to the existence of the ground effect (GE), which happens as the drones fly closely to the transmitting coils.
    As a basic study toward in-flight power transfer for drones, the focus of this paper is to develop suitable estimation algorithm.
    Although the ultrasonic sensor has sufficient ability for altitude measurement near the ground, angle measurement is not precise enough for advanced motion control\cite{Chandrasegar2023-mj}.
    For more precise estimation, the sensor fusion mainly combined with the inertial measurement unit (IMU) is utilized\cite{Li2021-bg}.
    In \cite{Svacha2019-oq}, Svacha et al. focused on the motor speed as the additional information for the proposed model-based sensor fusion to enhance the performance; however, the GE was not addressed by the aforementioned studies.
    \par
    With respect to the above discussion, this paper shows that by properly addressing the GE, the motor current can be effectively utilized to estimate the drone attitude via a model-based algorithm.
    Previous researches only mentioned the infinite model between the motor current and the altitude in the GE region, which shows no current is flowing when the altitude is zero.
    This paper proposes a new finite motor current model which can show a non-zero current in the GE region based on \cite{He2017-ja,He2019-yp,He2020-it}, which discusses the finite thrust model in the GE region.
    The model is identified experimentally.
    Then, attitude estimation with the proposed model is validated in the simulations and experiments.
    \par
    The reminder of this paper is organized as follows. In section $\mathrm{I\hspace{-1.2pt}I}$, the proposed model for attitude estimation from the motor equation is proposed. In section $\mathrm{I\hspace{-1.2pt}I\hspace{-1.2pt}I}$, the proposed method is verified in the simulations. In section $\mathrm{I\hspace{-1.2pt}V}$, the proposed method is verified in the experiments. In section $\mathrm{V}$, the conclusion is shown, and future studies are mentioned.

\section{Modeling and proposed attitude estimation}
    In this section, the attitude estimation method with the motor current model is proposed. First, previous studies about the in-GE motor power and thrust force model are introduced. Second, a motor current model for the GE region is proposed based on the motor equation and finite thrust model, and the attitude estimation relation for the altitude and the pitch angle is derived from the proposed motor current model. 
    Finally, the test bench dynamics is explained for the simulations and experimental evaluation of the proposed method.
    \subsection{Previous studies about motor power and thrust force model in GE region}
        In 1937, Betz proposed in \cite{Betz1937-ey} that the ratio of in-GE motor power to out-GE motor power is expressed as
        \begin{equation}
            \left[\frac{P}{P_{\infty}}\right]_{F=\mathrm { const.}}=\frac{2 z}{R},
            \label{eq:betz model}
        \end{equation}
        where $P$, $P_{\infty}$, $z$, and $R$ are the power of the drone, out-GE motor power of the drone, altitude of the propeller, and the propeller radius, respectively.
        This equation is only correct when $z \ll R$.
        In 1955, Cheeseman and Bennett proposed another expression of the aforementioned ratio in \cite{Cheeseman1955-ea} as
        \begin{equation}
            \left[\frac{P}{P_{\infty}}\right]_{F=\mathrm{const.}}=\frac{1}{1+\left(\frac{R}{4 z}\right)^2}.
        \end{equation}
        The above equation is derived from aerodynamic theories, such as blade element theory (BET).
        In 1976, Hayden proposed in \cite{Hayden1976-jh} that the ratio is expressed as
        \begin{equation}
            \left[\frac{P}{P_{\infty}}\right]_{F=\mathrm { const.}}=\frac{1}{A+B\left(\frac{2 R}{z}\right)^2},
            \label{eq:hayden's relation}
        \end{equation}
        where $A$ and $B$ are the function coefficients, respectively, which are determined experimentally.
        These models assume that the required power is zero on the ground.
        However, this assumption is not suitable to describe the drone motion at the ground level.
        Even if the altitude is zero, the drones need the thrust force to fly.
        Consequently, a certain power is required to rotate the propeller.
        The in-GE motor current model, which shows a non-zero value on the ground, is needed to estimate attitude from the model precisely.
        Unfortunately, no model has been proposed to satisfy such purpose.        
        \par
        The same situations occurred in the field for the modeling of in-GE thrust; the estimated thrust value surges with the previous in-GE thrust model, so many researchers tackled the finite thrust model.
        \par
        In that situation, the finite in-GE thrust model is proposed in \cite{He2017-ja,He2019-yp,He2020-it} by He et al. as
        \begin{equation}
            \left[\frac{F}{F_{\infty}}\right]_{P=\text { const. }}=1+C_a e^{-C_b z / R},
            \label{eq:he's thrust model}
        \end{equation}
        where the maximum ratio of thrust $C_a$ is calculated as
        \begin{equation}
            C_a=\frac{\sqrt{192 C_{l_\alpha} \sigma \theta_0+9 C_{l_\alpha} \sigma^2}-3 C_{l_\alpha} \sigma}{32 \theta_0+3 C_{l_\alpha} \sigma-\sqrt{192 C_{l_\alpha} \sigma \theta_0+9 C_{l_\alpha} \sigma^2}},
        \end{equation}
        where $C_{l_{\alpha}}$, $\sigma$, and $\theta_0$ are the lift coefficient, solidity of the propeller, and collective pitch angle of the propeller, respectively.
        $C_b$ shows the profile of the in-GE induced velocity.
        It can be calculated based on aerodynamic theories, such as BET.
        In the paper, $C_a$ is calculated from the propeller geometry, and $C_b$ is experimentally determined.
        In this paper, the model parameters are experimentally validated.
        Using this finite model, the in-GE motor current model, which shows a non-zero value on the ground, is derived in the next part.
        \begin{figure}[tb]
            \centering
            \includegraphics[width=0.85\linewidth]{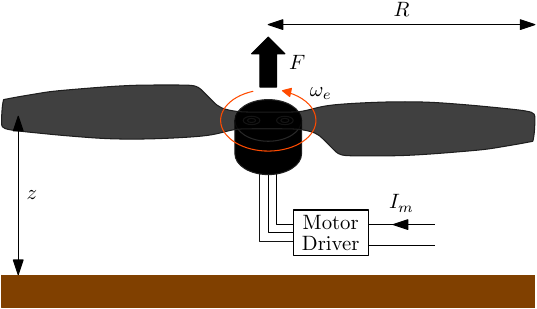}
            \caption{Propeller operational diagram in ground region.}
            \label{fig:propeller operational diagram in the ground effect region}
        \end{figure}
        
    \subsection{Proposed finite motor current model with respect to ground effect region}

        In this part, the finite motor current model is derived, assuming the constant battery voltage.
        The motor equation is written as
        \begin{equation}
            J_\omega \frac{\mathrm{d}\omega}{\mathrm{d}t}+D_\omega \omega = {K_\tau} I_m - C_Q \omega^2 - T_C,
            \label{eq:motor equation}
        \end{equation}
        where $J_\omega$, $D_\omega$, $\omega$, $K_\tau$, $I_m$, $C_Q$, and $T_C$ are the motor inertia, motor viscosity, rotational speed, torque coefficient, motor current, counter torque coefficient, and coulomb torque, respectively.
        When the motor is operating at the constant rotational speed $\omega_c$, the below equations are established:
        \begin{align}
            {K_\tau} I_m &= C_Q \omega_c^2+ D_\omega \omega_c+ T_C \approx C_Q\omega_c^2, \nonumber \\
            \therefore \hspace{3mm} I_m&=\frac{C_Q}{K_\tau} \omega_c^2.
            \label{eq:state value of motor current when hovering}
        \end{align}
        The approximation is validated because $D_\omega \omega_c, T_C \ll C_Q \omega_c^2$.
        \par
        When the drone is flying out-GE region, the thrust force needed for the flight $F$ is equal to the calculated output of the propeller force $F_e$ ($F=F_e$).
        Based on BET when the propeller rotational speed is $\omega_e$, the thrust force is expressed as
        \begin{equation}
            F_e=C_F \omega_e^2,
            \label{eq:BET equation when hovering OGE}
        \end{equation}
        where $C_F$ is the thrust coefficient.
        From \cref{eq:state value of motor current when hovering},
        \begin{equation}
            I_{m_{\infty}}=\frac{C_Q}{K_\tau} \omega_e^2
            \label{eq:OGE state value of motor current when hovering}
        \end{equation}
        is derived, where $I_{m_{\infty}}$ is the out-GE motor current.
        From \cref{eq:BET equation when hovering OGE,eq:OGE state value of motor current when hovering} and $F=F_e$,
        \begin{equation}
            I_{m_{\infty}}=\frac{C_Q}{K_\tau C_F} F
            \label{eq:relation between motor current and thrust OGE}
        \end{equation}
        is calculated.
        \par
\begin{figure}[tb]
            \centering
            \includegraphics[width=0.5\linewidth]{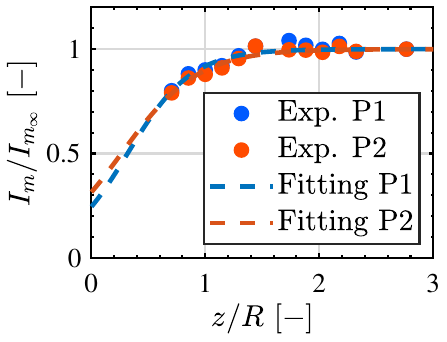}
            \caption{Identification result of the proposed motor current model.}
            \label{fig:identified proposed motor current model}
        \end{figure}

        When the drone is flying in the GE region,
        \begin{equation}
            F = C_{IGE} F_e
            \label{eq:force relation when hovering IGE}
        \end{equation}
        is derived, where $C_{IGE}$ is the in-GE thrust coefficient, and it is written as below based on \cref{eq:he's thrust model} in this paper:
        \begin{equation}
            C_{IGE} = 1+C_a e^{-C_b z / R}.
            \label{eq:thrust coefficient for hovering IGE}
        \end{equation}
        From \cref{eq:state value of motor current when hovering}, \cref{eq:BET equation when hovering OGE}, and \cref{eq:force relation when hovering IGE},
        \begin{equation}
            I_m=\frac{C_Q}{K_\tau C_F} \cdot \frac{F}{C_{I G E}}
            \label{eq:relation between motor current and thrust IGE}
        \end{equation}
        is calculated.
        Finally, from \cref{eq:relation between motor current and thrust OGE,eq:relation between motor current and thrust IGE},
        \begin{equation}
            \left[\frac{I_m}{I_{m_{\infty}}}\right]_{F=\text { const. }}=\frac{1}{C_{I G E}} =\frac{1}{1+C_a e^{-C_b z / R}}
            \label{eq:motor current model for altitude estimation}
        \end{equation}
        is derived.
        The coefficient of \cref{eq:motor current model for altitude estimation} is experimentally identified for each propeller as shown in \cref{fig:identified proposed motor current model}.
        The identified values are shown in \cref{table:simulation parameters}.
        The direct current component of the measured motor current is extracted by Fourier fast transform, and the coefficients are adjusted by least squares method.

\begin{figure}[tb]
            \centering
            \includegraphics[width=0.7\linewidth]{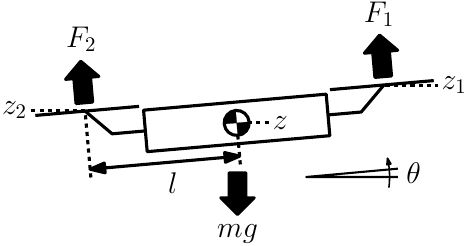}
            \caption{Two-degree-of-freedom drone model.}
            \label{fig:plant model of 2 DoF drone}
        \end{figure}

        From \cref{eq:motor current model for altitude estimation}, the altitude at the propeller is expressed with the motor current as
        \begin{equation}
            z = g(I_m) = -\frac{R}{C_b} \log \left\{\frac{1}{C_a}\left(\left[\frac{I_{m_{\infty}}}{I_m}\right]_{F=\text { const. }}-1\right)\right\}.
            \label{eq:proposed altitude relation}
        \end{equation}
        In this paper, the proposed attitude estimation method is implemented for the bench system of two-degree-of-freedom drone as shown in \cref{fig:plant model of 2 DoF drone}.
        Based on \cref{eq:proposed altitude relation}, each propeller altitude $z_1$ and $z_2$ is calculated, and the altitude of the gravitational point $z$ and the pitch angle $\theta$ as shown in \cref{fig:plant model of 2 DoF drone} are estimated as
        \begin{equation}
            z = \frac{z_1+z_2}{2}, \hspace{5mm} \theta = h\left(z_1, z_2\right) = \arcsin{\left( \frac{z_1-z_2}{2l} \right)}.
            \label{eq:estimated attitude}
        \end{equation}
        In the experiments, nonlinear recursive least squares (RLS) \cite{Umayahara2003-rd} is implemented to prevent the estimation deterioration due to the fluctuation of the measured motor current.
        
        \begin{figure*}[tb]
            \centering
            \includegraphics[width=1.0\linewidth]{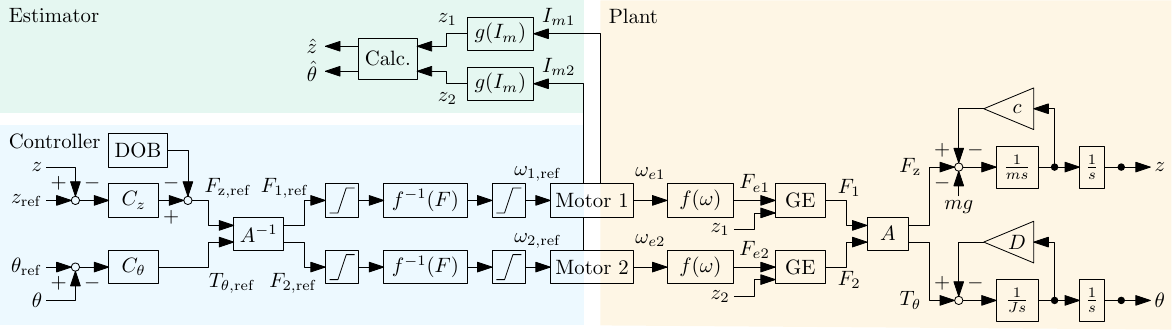}
            \caption{Block diagram of simulation system.}
            \label{fig:block diagram of simulation system}
        \end{figure*}

        \subsection{Test bench dynamics}
            The motion of the bench system in \cref{fig:plant model of 2 DoF drone} is described by  the following equations:
            \begin{subequations}
                \begin{align}
                    m\frac{\mathrm{d}^2z}{\mathrm{d}t^2} + c\frac{\mathrm{d}z}{\mathrm{d}t} &= F_1 \cos{\theta} + F_2 \cos{\theta} - mg = F_z - mg, \\
                    J\frac{\mathrm{d}^2\theta}{\mathrm{d}t^2} + D\frac{\mathrm{d}\theta}{\mathrm{d}t} &= F_1 l - F_2 l = T_{\theta},
                \end{align}
                \label{eq:equation of motion}
            \end{subequations}
            where $F_1$, $F_2$, $F_z$, $T_\theta$, $m$, $c$, $J$, $D$, $l$, and $g$ are the thrust force of propeller 1, thrust force of propeller 2, total thrust force affecting the body in parallel to the gravitational force, total torque affecting the body, total mass of the body, motor, and propeller, drag coefficient of the body, inertia of the body, rotational viscosity of the body, pitch moment arm, and gravitational constant, respectively.
            Besides, the below relations between $F_1$, $F_2$ and $F_z$, $T_{\theta}$ are established:
            \begin{subequations}
                \begin{align}
                    \left[\begin{array}{l}
                        F_z \\
                        T_\theta
                    \end{array}\right]&=\left[\begin{array}{cc}
                        \cos \theta & \cos \theta \\
                        l & -l
                    \end{array}\right]\left[\begin{array}{l}
                        F_1 \\
                        F_2
                    \end{array}\right] = A^{-1} \left[\begin{array}{l}
                        F_1 \\
                        F_2
                    \end{array}\right], \\
                    \left[\begin{array}{l}
                        F_1 \\
                        F_2
                    \end{array}\right]&=\left[\begin{array}{cc}
                        \frac{1}{2\cos{\theta}} & \frac{1}{2l} \\
                        \frac{1}{2\cos{\theta}} & -\frac{1}{2l}
                    \end{array}\right]\left[\begin{array}{l}
                        F_z \\
                        T_\theta
                    \end{array}\right] = A \left[\begin{array}{l}
                        F_z \\
                        T_\theta
                    \end{array}\right].
                \end{align}
            \end{subequations}
            Based on \cref{eq:equation of motion}, the nominal plant of the vertical motion and pitch motion can be defined as
            \begin{equation}
                G_z = \frac{1}{ms^2}, \hspace{5mm} G_J = \frac{1}{Js^2}.
                \label{eq:nominal plant}
            \end{equation}
            These nominal plants are utilized to design the controllers in the next section.

\section{Simulations}
    Simulations are conducted to verify the proposed method. 
    First, the simulation system is introduced based on the block diagram as shown in \cref{fig:block diagram of simulation system}.
    Second, attitude estimation with the proposed model is conducted in simulations.
    \par
    In \cref{fig:block diagram of simulation system}, $z_{\mathrm{ref}}$, $\theta_{\mathrm{ref}}$, $T_{z\mathrm{,ref}}$, $T_{\theta \mathrm{,ref}}$, $T_{1\mathrm{,ref}}$, $T_{2\mathrm{,ref}}$, $\omega_{1\mathrm{,ref}}$,and $\omega_{2\mathrm{,ref}}$ show the reference value of the altitude at the center of gravity (COG), pitch angle, thrust at the COG, moment around the COG, thrust at propeller 1, thrust at propeller 2, angular velocity at propeller 1, and angular velocity at propeller 2, respectively.
    $z$ and $\theta$ show the measured value of the altitude at the COG, and pitch angle, respectively.
   $\omega_{e1}$, $\omega_{e2}$, $F_{e1}$, and $F_{e2}$ show the rotational speed of propeller 1, rotational speed of propeller 2, thrust without GE at propeller 1, and thrust without GE at propeller 2, respectively.
    Function $f(\omega)$ is expressed as \cref{eq:BET equation when hovering OGE} and $f^{-1}(F)$ is expressed as the inverse of \cref{eq:BET equation when hovering OGE}.
    The GE is implemented as \cref{eq:he's thrust model}.
    $C_z$ and $C_{\theta}$ are designed as a PID and PD controller based on \cref{eq:nominal plant}, respectively. 
    Each controller is designed by multiple root pole placement, and each pole are set at $\SI{10}{rad/\second}$ and $\SI{30}{rad/\second}$, respectively.
    The motor model is implemented based on \cite{Naoki2023-jr}.
    The disturbance observer (DOB) is implemented for the altitude controller because the viscosity is large in the experimental bench system\cite{Nguyen2022-ew}.
    Other parameters in \cref{fig:block diagram of simulation system} are shown in \cref{table:simulation parameters}.

    \begin{table}[tb]
        \centering
        \caption{Simulation prameters.}
        \label{table:simulation parameters}
        \scalebox{0.85}{
            \begin{tabular}{ll} \toprule
                Parameter & Value \\ \midrule
                Torque Coefficient 1 $K_{\tau1}$ & $\SI{66.4}{m Nm/A}$ \\
                Torque Coefficient 2 $K_{\tau2}$ & $\SI{65.1}{m Nm/A}$ \\ 
                Inertia of Motor with Propeller 1 $J_{\omega 1}$ & $\SI{0.4}{kgm^2}$ \\
                Inertia of Motor with Propeller 2 $J_{\omega 2}$ & $\SI{0.392}{kgm^2}$ \\
                Motor Viscosity 1 $D_{\omega 1}$ & $\SI{4.6}{\micro ms/rad}$ \\
                Motor Viscosity 2 $D_{\omega 2}$ & $\SI{4.51}{\micro ms/rad}$ \\
                Counter Torque Coefficient 1 $C_{{Q 1}}$ & $\SI{9.56}{\micro Nms^2/rad^2}$ \\
                Counter Torque Coefficient 2 $C_{{Q 2}}$ & $\SI{9.88}{\micro Nms^2/rad^2}$ \\
                Coulomb Torque 1 $T_{{C 1}}$ & $\SI{2.4}{mN\metre}$ \\
                Coulomb Torque 2 $T_{{C 2}}$ & $\SI{2.35}{mN\metre}$ \\
                Total Mass of Body, Motor, and Propeller $m$ & $\SI{7.0}{\kilogram}$ \\
                Drag Coefficient of Body $c$ & $\SI{1.0}{m Ns/\metre}$ \\
                Inertia of Body $J$ & $\SI{2.34}{\kilogram \metre^2}$ \\
                Rotational Viscosity of Body $D$ & $\SI{0.1}{\micro Nms/rad}$ \\
                Pitch Moment Arm $l$ & $\SI{0.63}{\metre}$ \\
                Rotor Radius $R$ & $\SI{0.34}{\metre}$ \\
                Thrust Coefficient $C_F$ & $\SI{0.399}{m Ns^2/rad^2}$ \\
                Ground Effect Coefficient 1 $C_{{a\mathrm{P1}}}$ /  $C_{{a\mathrm{P2}}}$ & $3.11$ / $2.20$ \\
                Ground Effect Coefficient 2 $C_{{b\mathrm{P1}}}$ /  $C_{{b\mathrm{P2}}}$ & $3.56$ / $2.97$ \\ \bottomrule
            \end{tabular}        
        }
    \end{table}

        \begin{figure*}[tb]
            \centering
            \subfigure[]{
                  \includegraphics[width=0.23\linewidth]{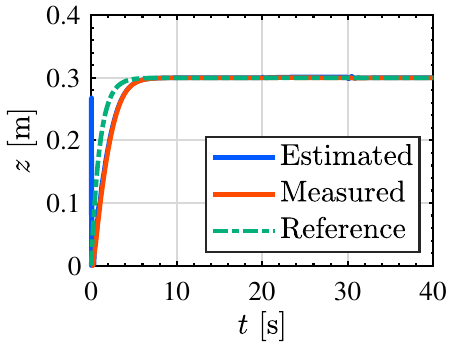}
                  \label{fig:simulation results of altitude z}}
            \subfigure[]{
                \includegraphics[width=0.23\linewidth]{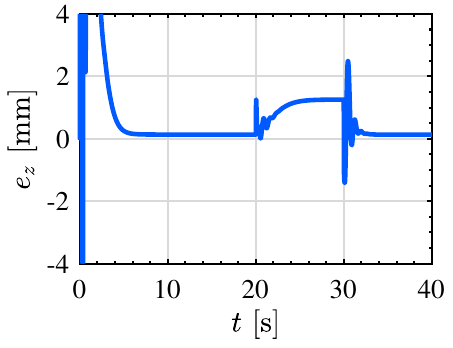}
                \label{fig:simulation results of error z}}
            \subfigure[]{
                \includegraphics[width=0.23\linewidth]{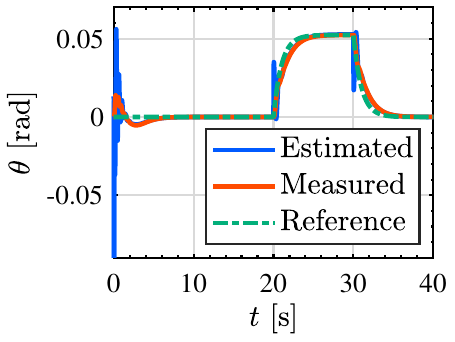}
                \label{fig:simulation results of angle theta}}
            \subfigure[]{
                \includegraphics[width=0.23\linewidth]{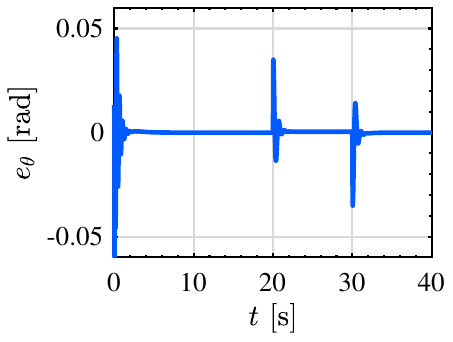}
                \label{fig:simulation results of error theta}}
            \subfigure[]{
                \includegraphics[width=0.23\linewidth]{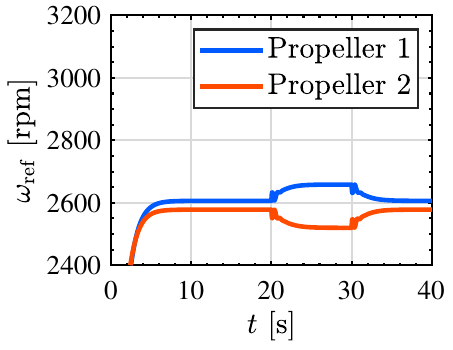}
                \label{fig:simulation results of omega}}
            \subfigure[]{
                  \includegraphics[width=0.23\linewidth]{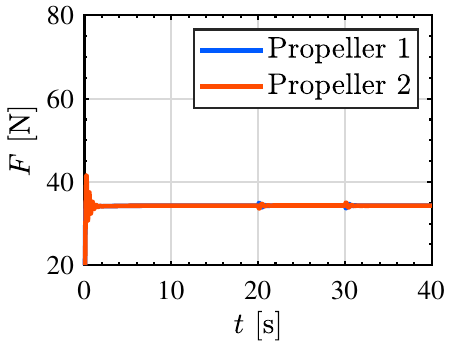}
                  \label{fig:simulation results of thrust}}
            \subfigure[]{
                \includegraphics[width=0.23\linewidth]{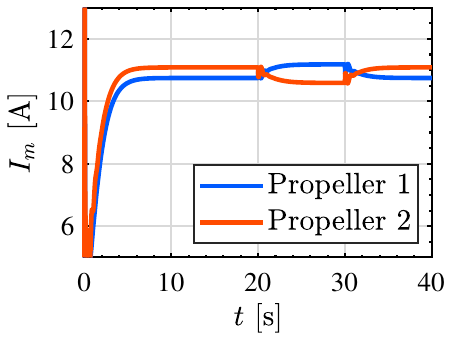}
                \label{fig:simulation results of current}}
            \caption{Simulation results of altitude and pitch angle estimation. (a) Altitude $z$ (b) Altitude estimation error $e_z$ (c) Pitch angle $\theta$ (d) Pitch Angle estimation error $e_\theta$ (e) Angular velocity of motor $\omega$ (f) Thrust $F$ (g) Motor current $I_m$}
            \label{fig:simulation results for altitude estimation with proposed model}
        \end{figure*}
    
        Simulation results to verify the attitude estimation method with the proposed model are shown in \cref{fig:simulation results for altitude estimation with proposed model}.
        The simulations are conducted assuming in-flight inductive power transfer system.
        In the simulations, the drone starts to fly at \SI{0}{\second} and hovers at $\SI{0.3}{\metre}$ height which is in the GE region.
        At $\SI{20}{\second}$, the pitch angle rises to \SI{0.05}{rad}, which demonstrates the forward flight, and goes down to \SI{0.0}{rad} while maintaining the same height.
        In this condition, the distance between the transfer coil and receiver coil is assumed to be $\SI{0.1}{\metre}$, and the coupling coefficient of the system, which means that the coupling strength between the transfer coil and receiver coil is assumed to be 0.10, which is the reasonable value for the efficient inductive power transfer.
        \par
        \cref{fig:simulation results of altitude z,fig:simulation results of error z} show the altitude responses.
        In the state condition, the maximum error of the altitude estimation is below $\SI{3}{m\metre}$, which is sufficient for the efficient in-flight inductive power transfer.
        \cref{fig:simulation results of angle theta,fig:simulation results of error theta} show the pitch angle responses.
        In the state condition, the maximum error of the pitch angle estimation is below $\SI{0.04}{rad}$, which is also sufficient to the in-flight inductive power transfer.
        \cref{fig:simulation results of omega,fig:simulation results of thrust,fig:simulation results of current} show the motor rotational speed response, thrust response, and motor current response, respectively.
        In these figures, it is observed that the differences are increased when the reference value of the pitch angle is non-zero. 
        It can be seen that the state values in \cref{fig:simulation results of omega,fig:simulation results of thrust,fig:simulation results of current} do not coincide because the motor parameters are not the same in the simulations.
        However, the differences are normalized with the out-GE motor current $I_{m_\infty}$ shown in \cref{fig:identified proposed motor current model}
        Therefore, the motor parameters deviations do not affect the estimation performance in the simulations.
        \par
        The simulation results with model parameter errors and the same reference input of the attitude are shown in \cref{table:evaluation results of the parameter errors}.
        The estimation errors and root mean square deviation (RMSD) are shown in \cref{table:evaluation results of the parameter errors}.
        $C_{an}$ and $C_{bn}$ are the nominal model parameters.
        It is assumed that the nominal parameters deviate from the actual coefficients $C_{a}$ and $C_{b}$.
        From \cref{table:evaluation results of the parameter errors}, it is observed that the maximum errors of the altitude and the pitch angle are $\SI{+23}{m\metre}$ and $\SI{\pm 0.04}{rad/s}$.
        
        \begin{table}[tb]
            \centering
            \caption{Evaluation results of parameter errors.}
            \label{table:evaluation results of the parameter errors}
            \scalebox{0.7}{
                \begin{tabular}{lrr} \toprule
                     & Altitude error / RMSD & Pitch angle error / RMSD \\ \midrule
                    Without Model Errors & $\SI{2}{m\metre}$ / $\SI{10}{m\metre}$ & $\SI{\pm 0.04}{rad}$ / $\SI{6.79}{m rad}$ \\
                    $C_{an}=1.05 C_a, C_{bn}=1.05 C_b$ & $\SI{-10}{m\metre}$ / $\SI{12.9}{m\metre}$ & $\SI{\pm 0.04}{rad}$ / $\SI{6.69}{m rad}$ \\
                    $C_{an}=1.05 C_a, C_{bn}=0.95 C_b$ & $\SI{+23}{m\metre}$ / $\SI{24.1}{m\metre}$ & $\SI{\pm 0.04}{rad}$ / $\SI{7.30}{m rad}$ \\
                    $C_{an}=0.95 C_a, C_{bn}=1.05 C_b$ & $\SI{-20}{m\metre}$ / $\SI{20.6}{m\metre}$ & $\SI{\pm 0.04}{rad}$ / $\SI{6.49}{m rad}$ \\
                    $C_{an}=0.95 C_a, C_{bn}=0.95 C_b$ & $\SI{+12}{m\metre}$ / $\SI{14.2}{m\metre}$ & $\SI{\pm 0.04}{rad}$ / $\SI{7.39}{m rad}$ \\ \bottomrule
                \end{tabular}
            }
        \end{table}
        
\section{Experiments}

\begin{figure}[tb]
        \centering
        \includegraphics[width=0.9\linewidth]{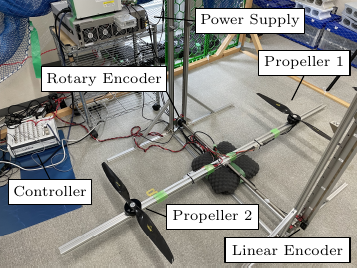}
        \caption{Drone bench system for experimental validation.}
        \label{fig:drone bench system for experimental validation}
    \end{figure}

    In this section, the experimental validation is conducted in the bench system as shown in \cref{fig:drone bench system for experimental validation}.
    The bench system consists of the two propellers, linear encoder, rotational encoder, controller, and power supply.
    The controller for the bench system is the same for the simulations as shown in \cref{fig:block diagram of simulation system}.

    \begin{figure*}[tb]
        \centering
        \subfigure[]{
              \includegraphics[width=0.23\linewidth]{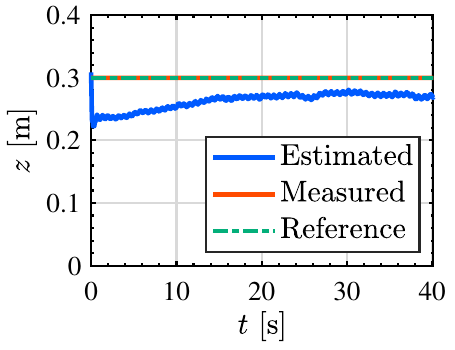}
              \label{fig:experimental results of altitude z}}
        \subfigure[]{
            \includegraphics[width=0.23\linewidth]{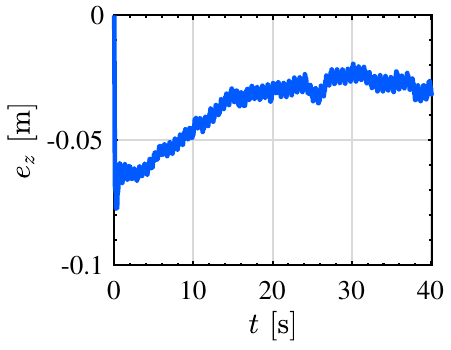}
            \label{fig:experimental results of error z}}
        \subfigure[]{
            \includegraphics[width=0.23\linewidth]{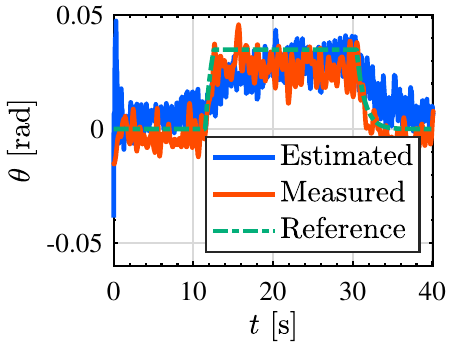}
            \label{fig:experimental results of angle theta}}
        \subfigure[]{
            \includegraphics[width=0.23\linewidth]{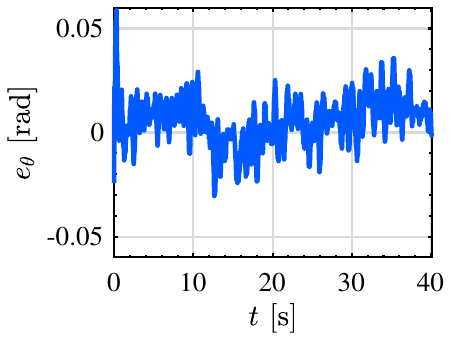}
            \label{fig:experimental results of error theta}}
        \subfigure[]{
            \includegraphics[width=0.23\linewidth]{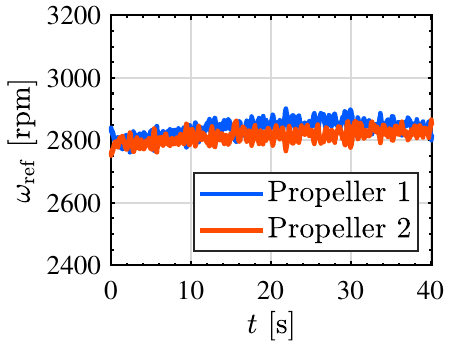}
            \label{fig:experimental results of omega}}
        \subfigure[]{
              \includegraphics[width=0.23\linewidth]{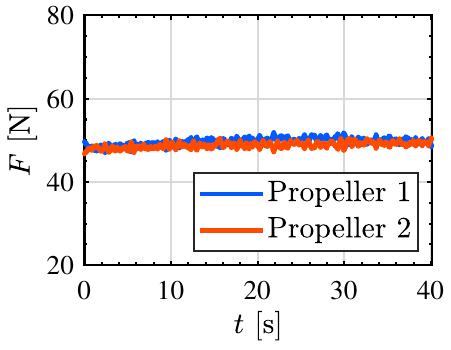}
              \label{fig:experimental results of thrust}}
        \subfigure[]{
            \includegraphics[width=0.23\linewidth]{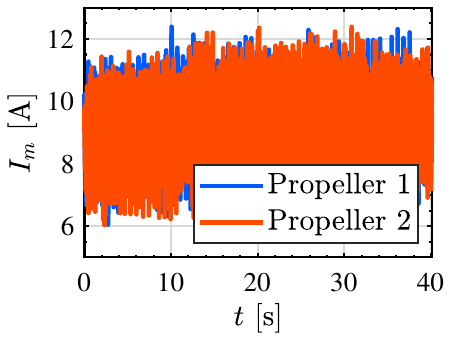}
            \label{fig:experimental results of current}}
        \subfigure[]{
            \includegraphics[width=0.23\linewidth]{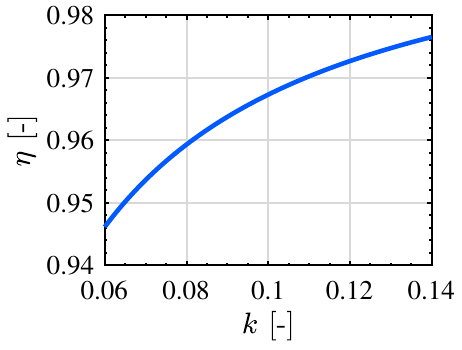}
            \label{fig:transmission efficiency of the system}}
        \caption{Experimental results for altitude and pitch angle estimation. (a) Altitude $z$ (b) Altitude estimation error $e_z$ (c) Pitch angle $\theta$ (d) Pitch Angle estimation error $e_\theta$ (e) Angular velocity of motor $\omega$ (f) Thrust $F$ (g) Motor current $I_m$ (h) Transmission efficiency of the inductive power transfer system at $\omega=\SI{85}{kHz}$, $R_1=\SI{108}{m\Omega}$, $R_2=\SI{32.5}{m\Omega}$, $L_1=\SI{236}{\micro H}$, $L_2=\SI{18.9}{\micro H}$ on the resonance condition $\eta$}
        \label{fig:experimental results for altitude estimation with proposed model}
    \end{figure*}
    
    \par
    Experimental results are shown in \cref{fig:experimental results for altitude estimation with proposed model}.
    \cref{fig:experimental results of omega,fig:experimental results of thrust,fig:experimental results of current} show the reference value of the motor rotational speed, thrust, and motor current, respectively.
    There exist signal fluctuations due to motor vibration and sensor noises.

    
    \par
    \cref{fig:experimental results of altitude z,fig:experimental results of error z} show the altitude responses.
    The estimations are conducted based on the fluctuating motor currents shown in \cref{fig:experimental results of current}, so nonlinear RLS mentioned before is implemented.
    Its forgetting factor is 0.9985 for each propeller.
    From \cref{fig:experimental results of error z}, the state maximum error of the altitude estimation from the measured altitude value is $\SI{-0.03}{\metre}$ and RMSD of the error is $\SI{0.0391}{\metre}$.
    When the drone flies at $\SI{0.30}{\metre}$, the coupling coefficient $k$ and the transmission efficiency of the inductive power transfer system $\eta$ as shown in \cref{fig:transmission efficiency of the system} is 0.10 and $\SI{96.7}{\percent}$, respectively.
    If there is an altitude deviation of $\SI{0.03}{\metre}$, the coupling coefficient varies between 0.07 and 0.13\cite{Fujimoto2021-jd}.
    This deviation of the coupling coefficient leads to the deviation of the transmission efficiency from $\SI{95.3}{\percent}$ to $\SI{97.5}{\percent}$, which is sufficient for the efficient inductive power transfer.
    As shown above, the estimation has the enough performance for the efficient inductive power transfer.
    \cref{fig:experimental results of angle theta} and \cref{fig:experimental results of error theta} show the pitch angle responses.
    In the state condition, the maximum error of the pitch angle estimation is below $\SI{0.04}{rad}$ and RMSD of the error is $\SI{0.0130}{rad}$, which is also sufficient for the in-flight inductive power transfer.

\section{Conclusion}
    In this paper, we proposed a novel motor current model based on the motor equation at the constant rotational speed and finite thrust model.
    It is shown that the proposed model can fit the measured results, and estimation performance of the model is validated with the simulations and experiments.
    The estimation errors are $\SI{0.30}{\metre}$ maximum with respect to the altitude, and $\SI{0.04}{rad}$ maximum with respect to the pitch angle in the experiments.
    These estimation errors might lead to the efficiency deviation from $\SI{95.3}{\percent}$ to $\SI{97.5}{\percent}$, which is enough for the efficient inductive power transfer.
    \par
    In the future, the proposed method will be extended by considering the fusion of the current sensor with the other motion sensors, such as the IMU, ultrasonic sensor, and on-board vision system.
    In parallel, the robustness of the proposed method for the model parameter errors is investigated.
    The estimated attitude will be applied for the precise flight control scheme on the in-flight inductive power transfer system.
    The demonstration of the in-flight inductive power transfer with a real vehicle is also conducted as the demonstration of the precise flight control near the ground.

\section*{Acknowledgement}
    This work was partly supported by JSPS KAKENHI Grant Number JP23H00175, Japan.

\bibliographystyle{IEEEtran}
\bibliography{list.bib}

\end{document}